\documentclass[10pt,twocolumn,letterpaper]{article}

\usepackage[pagenumbers]{cvpr} 
\usepackage{multirow}
\usepackage{url}
\usepackage{wrapfig}
\usepackage{bm}
\usepackage{colortbl}
\usepackage{listings}
\usepackage[dvipsnames]{xcolor}

\definecolor{cvprblue}{rgb}{0.21,0.49,0.74}
\definecolor{dark-gray}{gray}{0.20}
\definecolor{mygreen}{HTML}{39b54a}

\newcommand{\pub}[1]{{\color{dark-gray}{\tiny{[{#1}]}}}}
\usepackage[pagebackref,breaklinks,colorlinks,citecolor=cvprblue]{hyperref}

\def\paperID{*****} 
\def\confName{CVPR}
\def\confYear{2024}

\title{UniRepLKNet: A Universal Perception Large-Kernel ConvNet for Audio, Video, Point Cloud, Time-Series and Image Recognition}

\author{
	{Xiaohan Ding}$^{1}$\thanks{Equal contributions.  }
	~~~ {Yiyuan Zhang}$^{2\ast}$
	~~~ {Yixiao Ge}$^{1}$\\
	~~~ {Sijie Zhao}$^{1}$
        \quad \quad~~ {Lin Song}$^{1}$
        \quad \quad~~ {Xiangyu Yue}$^{2}$
	\quad \quad~~ {Ying Shan}$^{1}$ \\
	\textsuperscript{1} Tencent AI Lab
        \quad 
	~~\textsuperscript{2}~ The Chinese University of Hong Kong\\
        \texttt{\url{https://github.com/AILab-CVC/UniRepLKNet}}
}

\begin{document}
\maketitle
\begin{abstract}
Large-kernel convolutional neural networks (ConvNets) have recently received extensive research attention, but two unresolved and critical issues demand further investigation. 1) The architectures of existing large-kernel ConvNets largely follow the design principles of conventional ConvNets or transformers, while the architectural design for large-kernel ConvNets remains under-addressed. 2) As transformers have dominated multiple modalities, it remains to be investigated whether ConvNets also have a strong universal perception ability in domains beyond vision. In this paper, we contribute from two aspects. 1) We propose four architectural guidelines for designing large-kernel ConvNets, the core of which is to exploit the essential characteristics of large kernels that distinguish them from small kernels - they can see wide without going deep. Following such guidelines, our proposed large-kernel ConvNet shows leading performance in image recognition (ImageNet accuracy of 88.0\%, ADE20K mIoU of 55.6\%, and COCO box AP of 56.4\%), demonstrating better performance and higher speed than the recent powerful competitors. 2) We discover large kernels are the key to unlocking the exceptional performance of ConvNets in domains where they were originally not proficient. With certain modality-related preprocessing approaches, the proposed model achieves state-of-the-art performance on time-series forecasting and audio recognition tasks even without modality-specific customization to the architecture. All the code and models are publicly available on GitHub and Huggingface.
\end{abstract}    
\section{Introduction}
\label{sec:intro}

The design paradigm of convolutional neural networks (ConvNets) with very large kernels originated from RepLKNet~\cite{ding2022scaling} when the status of ConvNets was challenged by Vision Transformers (ViTs)~\cite{vit,swin,deit,pvt,ge2023advancing}. Inspired by ViTs that use global attention~\cite{vit,pvt,bot} or attention with large windows~\cite{swin,halonet,sasa}, RepLKNet proposed to use very large conv kernels. In contrast to the common practice using small kernels (\eg, 3$\times$3)~\cite{simonyan2014very,he2016deep,zhang2018shufflenet,mbv1,huang2017densely,efficientnet,regnet}, which fails to obtain a large Effective Receptive Field (ERF)~\cite{erf} even with numerous small-kernel layers, RepLKNet realizes large ERF and impressive performance, especially on tasks such as object detection and semantic segmentation. 

Nowadays, ConvNets with very large kernels become popular, which mostly focus on making the large kernels even larger~\cite{liu2022more}, ways to apply them to multiple tasks~\cite{chen2023largekernel3d,luo2023lkd,xie2023large}, \etc. However, we note that most architectures of the existing large-kernel ConvNets simply follow other models, \eg, RepLKNet~\cite{ding2022scaling} follows the architecture of Swin Transformer~\cite{liu2021swin}, and SLaK~\cite{liu2022more} follows ConvNeXt, which is a powerful architecture with medium-sized (7$\times$7) kernels. The architectural design for large-kernel ConvNets remains under-explored.

We explore large-kernel ConvNet architecture by rethinking the design of conventional models that employ a deep stack of small kernels. As we add a 3$\times$3 conv to a small-kernel ConvNet, we expect it to take three effects simultaneously - \textbf{1)} make the receptive field larger, \textbf{2)} increase the abstract hierarchy of spatial patterns (\eg, from angles and textures to shapes of objects), and \textbf{3)} improve the model's general representational capability via making it deeper, bringing in more learnable parameters and non-linearities. In contrast, we argue that such three effects in a large-kernel architecture should be decoupled as the model should utilize the substantial strength of a large kernel - \emph{the ability to see wide without going deep}. Since increasing the kernel size is much more effective than stacking more layers in enlarging the ERF~\cite{erf}, a sufficient ERF can be built up with a small number of large-kernel layers, so that the compute budget can be saved for other efficient structures that are more effective in increasing the abstract hierarchy of spatial patterns or generally increasing the depth. For example, when the objective is to extract higher-level local spatial patterns from lower-level ones, a 3$\times$3 might be a more suitable option than a large-kernel conv layer. The reason is that the latter demands more computations and may result in patterns no longer restricted to smaller local regions, which could be undesirable in specific scenarios.

\begin{figure}
    \begin{center}
            \includegraphics[width=0.93\linewidth]{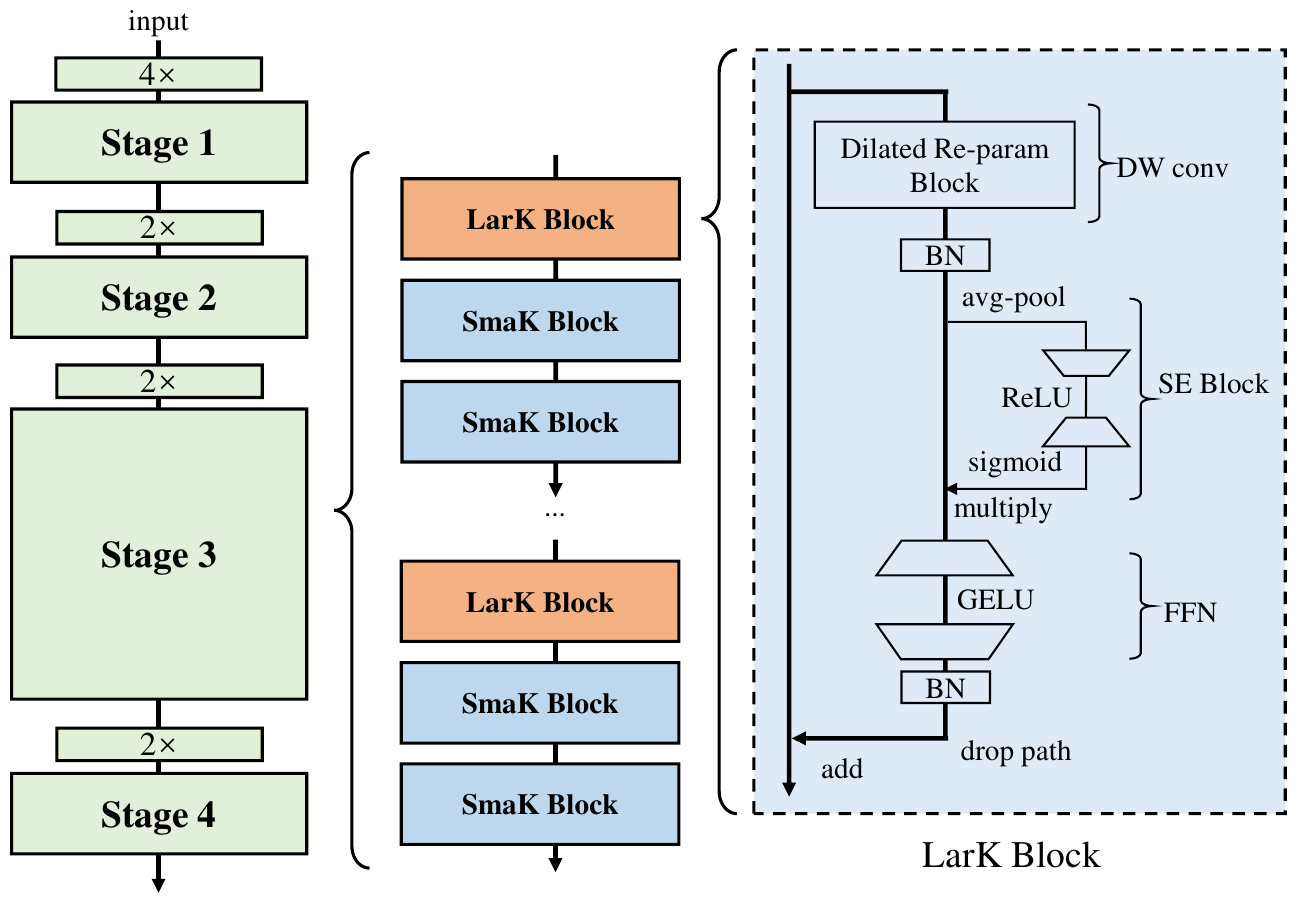}
        \vspace{-2mm}
        \caption{Architectural design of UniRepLKNet. A LarK Block comprises a Dilated Reparam Block proposed in this paper, an SE Block~\cite{hu2018squeeze}, an FFN, and Batch Normalization (BN)~\cite{ioffe2015batch} layers. The only difference between a SmaK Block and a LarK Block is that the former uses a depth-wise 3$\times$3 conv layer in replacement of the Dilated Reparam Block in the latter. Stages are connected by downsampling blocks implemented by stride-2 dense 3$\times$3 conv layers. We may flexibly arrange the blocks in different stages and the details of our provided instances are shown in Table~\ref{table-instances}.}
        \label{fig-arch}
        \vspace{-0.3in}
    \end{center}
\end{figure}

Concretely, we propose \textbf{four architectural guidelines} for large-kernel ConvNets - \textbf{1)} use efficient structures such as SE Blocks~\cite{hu2018squeeze} to increase the depth, \textbf{2)} use a proposed \emph{Dilated Reparam Block} to re-parameterize the large-kernel conv layer to \emph{improve the performance without inference costs}, \textbf{3)} decide the kernel size by the downstream task and usually use large kernels only in the middle- and high-level layers, and \textbf{4)} add 3$\times$3 conv instead of more large kernels while scaling up the model's depth. A ConvNet built up following such guidelines (Fig.~\ref{fig-arch}) realizes the aforementioned three effects separately, as it uses a modest number of large kernels to guarantee a large ERF, small kernels to extract more complicated spatial patterns more efficiently, and multiple lightweight blocks to further increase the depth to enhance the representational capacity.

Our architecture achieves leading performance on ImageNet classification~\cite{deng2009imagenet}, ADE20K semantic segmentation~\cite{zhou2019semantic}, and COCO object detection~\cite{lin2014microsoft}, outperforming the existing large-kernel ConvNets such as RepLKNet~\cite{ding2022scaling}, SLaK~\cite{liu2022more}, and recent powerful architectures including ConvNeXt V2~\cite{woo2023convnext}, FastViT~\cite{vasu2023fastvit}, Swin V2~\cite{liu2022swin} and DeiT III~\cite{touvron2022deit} in terms of both accuracy and efficiency. Moreover, our architecture demonstrates significantly higher shape bias~\cite{tuli2021convolutional,modelvshuman} than existing ConvNets and ViTs, \ie, it makes predictions more based on the overall shapes of objects than the textures, which agrees with the human visual system and results in better generalization. This may explain its superiority in downstream tasks. See the Appendix for details.

RepLKNet~\cite{ding2022scaling} was proposed partly ``in defense of ConvNets'' as ViTs dominated multiple image recognition tasks that were once dominated by ConvNets. Moreover, considering transformers have shown universal perception capability in multiple modalities~\cite{zhang2024multimodal,zhang2023meta}, in this work, we seek to not only reclaim the leading position in image recognition tasks by surpassing ViTs' performance but also contribute to areas where ConvNets were not traditionally dominant. Specifically, on audio, video, point cloud, and time-series tasks, we achieve impressive performance with amazingly universal and simple solutions. We use modality-specific preprocessing approaches to transform all the data into 3D embedding maps just like what we do with images and use the same architecture as the backbone to process the embedding maps. Our model shows \textbf{\emph{uni}versal perception ability across multiple modalities with a \emph{uni}fied architecture} so it is named \textbf{UniRepLKNet}. 

Impressively, UniRepLKNet achieves remarkable results even on modalities that were not considered the stronghold of ConvNet, \eg, audio and temporal data. On a huge-scale time-series forecasting task that predicts the global temperature and wind speed, UniRepLKNet, a generalist model originally designed for image recognition, even outperforms the latest state-of-the-art transformer customized for the task. Such results not only signify a \emph{``\textbf{comeback}''} for ConvNet in its original domain but also showcase large-kernel ConvNet's potential to \emph{``\textbf{conquer}''} new territories, expanding its applicability and versatility in various tasks.

\section{Related Work}

\noindent\textbf{Large kernels in early ConvNets}. Classic ConvNets such as AlexNet~\cite{krizhevsky2012imagenet} and Inceptions~\cite{szegedy2015going,szegedy2016rethinking,szegedy2017inception} used 7$\times$7 or 11$\times$11 in the low-level layers, but large kernels became not popular after VGG-Net~\cite{simonyan2014very}. Global Convolution Network (GCN)~\cite{peng2017large} used very large conv layers (1$\times$K followed by K$\times$1) for semantic segmentation. Local Relation Networks (LR-Net)~\cite{hu2019local} adopted a spatial aggregation operator (LR-Layer) to replace the standard conv layer, which can be viewed as a dynamic convolution. LR-Net benefited from a kernel size of 7$\times$7 but degraded with 9$\times$9. With a kernel size as large as the feature map, its top-1 accuracy significantly reduced from 75.7\% to 68.4\%. 

\noindent\textbf{Explorations with large kernels}. The concept of kernel may be generalized beyond spatial convolution. Swin Transformer~\cite{swin} used shifted attention with window sizes ranging from 7 to 12, which can be seen as a dynamic kernel. \citet{han2021demystifying} replaced the attention layers in Swin with static or dynamic 7$\times$7 conv and still maintained comparable results. MetaFormer~\cite{yu2021metaformer} suggested large-kernel pooling layer was an alternative to self-attention. Another representative work was Global Filter Network (GFNet)~\cite{rao2021global}, which optimized the spatial connection weights in the Fourier domain. It is equivalent to circular global convolutions in the spatial domain. 

\noindent\textbf{Modern ConvNets with very large kernels}. RepLKNet first proposed that simply scaling up the kernel size of existing ConvNets resulted in improvements, especially on downstream tasks~\cite{ding2022scaling}. It proposed several guidelines while using large kernels, which were focused on the microstructural design (\eg, using shortcut alongside large kernel) and application (large-kernel ConvNets should be evaluated on downstream tasks). In terms of the architecture, RepLKNet merely followed Swin Transformer for simplicity. In the past two years, large-kernel ConvNets have been intensively studied. Some works succeeded in further enlarging the kernel sizes~\cite{liu2022more}, generalizing the idea to 3D scenarios~\cite{chen2023largekernel3d} and many downstream tasks, \eg, image dehazing~\cite{luo2023lkd} and super-resolution~\cite{xie2023large}. However, we note that the architectural design for ConvNets with very large kernels remains under-explored. For example, SLaK~\cite{liu2022more} followed the architecture developed by ConvNeXt, which is a powerful architecture of medium-sized (7$\times$7) kernels.

\section{Architectural Design of UniRepLKNet}

We first summarize the architectural guidelines as follows. \textbf{1)} Block design: use efficient structures that perform both inter-channel communications and spatial aggregations to increase the depth. \textbf{2)} Re-parameterization: use dilated small kernels to re-parameterize a large kernel. \textbf{3)} Kernel size: decide kernel size according to the downstream task and usually use large kernels in middle- and high-level layers. \textbf{4)} Scaling Rule: while scaling up the depth, the added blocks should use small kernels. We describe the proposed Dilated Reparam Block in Sec.~\ref{sec-dil-reparam} and details in Sec.~\ref{sec-guidelines}.

\subsection{Dilated Reparam Block}~\label{sec-dil-reparam}

	\begin{figure*}
		\begin{center}
            \includegraphics[width=0.83\linewidth]{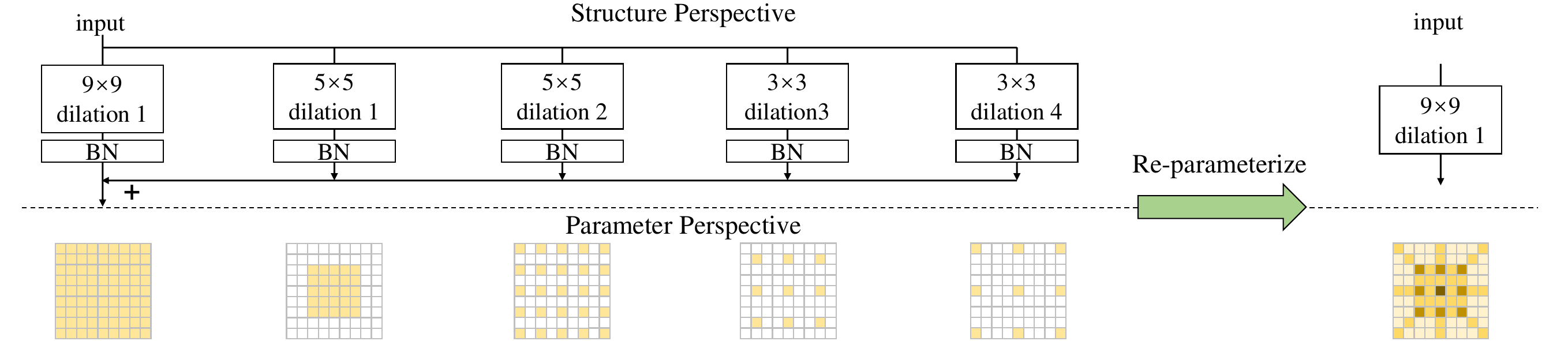}
            \vspace{-0.1in}
			\caption{Dilated Reparam Block uses dilated small-kernel conv layers to enhance a non-dilated large-kernel layer. Such dilated layers are equivalent to a non-dilated conv layer with a larger sparse kernel, as shown from the parameter perspective, so that the whole block can be equivalently transformed into a single large-kernel conv. This example shows $K$=9, and we may use more dilated layers for larger $K$.}
			\label{fig-reparam}
			\vspace{-0.25in}
		\end{center}
	\end{figure*}

It is reported a large-kernel conv should be used with a parallel small-kernel one because the latter helps capture the small-scale patterns during training~\cite{ding2022scaling}. Their outputs are added up after two respective Batch Normalization (BN)~\cite{ioffe2015batch} layers. After training, with the Structural Re-parameterization~\cite{ding2021repvgg,ding2019acnet,ding2021diverse,ding2021repmlpnet,ding2021resrep,ding2022re} methodology, we merge the BN layers into the conv layers so the small-kernel conv can be equivalently merged into the large-kernel one for inference. In this work, we note that except for small-scale patterns, enhancing the large kernel's capability to capture sparse patterns (\ie, a pixel on a feature map may be more related to some distant pixels than its neighbors) may yield features of higher quality. The need to capture such patterns exactly matches the mechanism of dilated convolution - from a sliding-window perspective, a dilated conv layer with a dilation rate of $r$ scans the input channel to capture spatial patterns where each pixel of interest is $r-1$ pixels away from its neighbor. Therefore, we use dilated conv layers parallel to the large kernel and add up their outputs.

To eliminate the inference costs of the extra dilated conv layers, we propose to equivalently transform the whole block into a single non-dilated conv layer for inference. Since \emph{ignoring pixels of the input is equivalent to inserting extra zero entries into the conv kernel}, \emph{a dilated conv layer with a small kernel can be equivalently converted into a non-dilated (\ie, $r=1$) layer with a sparse larger kernel}. Let $k$ be the kernel size of the dilated layer, by inserting zero entries, the kernel size of the corresponding non-dilated layer will be $(k-1)r+1$, which is referred to as the \emph{equivalent kernel size} for brevity. We further note that such transformation from the former kernel $\mathrm{W}\in\mathcal{R}^{k\times k}$ to the latter $\mathrm{W}^\prime\in\mathcal{R}^{((k-1)r+1)\times ((k-1)r+1)}$ can be elegantly realized by a transpose convolution with a stride of $r$ and an identity kernel $\mathrm{I}\in\mathcal{R}^{1\times1}$, which is scalar 1 but viewed as a kernel tensor.~\footnote{We showcase a single-channel conv and it is easy to generalize the transformation to multi-channel cases. See the Appendix for details.} With pytorch-style pseudo code, that is
\begin{equation}\label{eq-merge}
    \mathrm{W}^\prime = \mathtt{conv\_transpose2d}(\mathrm{W}, \mathrm{I}, \text{stride}=r) \,.
\end{equation}
The equivalency can be easily verified - given an arbitrary $\mathrm{W}\in\mathcal{R}^{k\times k}$ and an arbitrary input channel, a convolution with $\mathrm{W}$ and a dilation rate $r$ always yields identical results to a non-dilated convolution with $\mathrm{W}^\prime$.~\footnote{In common cases where the shape of output equals that of input, \ie, the padding of the former is $\frac{k-1}{2}$, note the padding of the latter should be $\frac{(k-1)r}{2}$ since the size of the equivalent sparse kernel is $(k-1)r+1$.}

Based on such equivalent transformations, we propose a Dilated Reparam Block, which uses a non-dilated small-kernel and multiple dilated small-kernel layers to enhance a non-dilated large-kernel conv layer. Its hyper-parameters include the size of large kernel $K$, sizes of parallel conv layers $\bm{k}$, and the dilation rates $\bm{r}$. The shown case (Fig.~\ref{fig-reparam}) with four parallel layers is denoted by $K$=9, $\bm{r}$=(1,2,3,4), $\bm{k}$=(5,5,3,3). For a larger $K$, we may use more dilated layers with larger kernel sizes or dilation rates. The kernel sizes and dilation rates of the parallel branches are flexible and the only constraint is $(k-1)r+1 \leq K$. For example, with $K$=13 (the default setting in our experiments), we use five layers with $\bm{k}$=(5,7,3,3,3), $\bm{r}$=(1,2,3,4,5), so the equivalent kernel sizes will be (5,13,7,9,11), respectively. To convert a Dialted Reparam Block into a large-kernel conv layer for inference, we first merge every BN into the preceding conv layer, convert every layer with dilation $r>1$ with function~\ref{eq-merge}, and add up all the resultant kernels with appropriate zero-paddings. For example, the layer in Fig.~\ref{fig-reparam} with $k$=3,$r$=3 is converted into a sparse 7$\times$7 kernel and added to the 9$\times$9 kernel with one-pixel zero paddings on each side.

\subsection{Architectural Guidelines for Large Kernels}\label{sec-guidelines}

\textbf{Vanilla architecture}. We first construct the vanilla architecture to experiment on. As a common practice, the main body of the model is split into four stages connected by downsampling blocks. Specifically, the first downsampling block uses two stride-2 3$\times$3 conv layers to transform the raw input into $C$-channel feature maps, where $C$ is an architectural hyper-parameter and the other three downsampling blocks each use one stride-2 3$\times$3 conv layer performing 2$\times$ channel expansion so that the numbers of channels of the four stages are $C$, $2C$, $4C$, and $8C$, respectively. A stage comprises blocks whose vanilla design resembles ConvNeXt, \ie, a \emph{depthwise (DW)} conv layer and a \emph{Feed-Forward Network (FFN)} with GRN unit~\cite{woo2023convnext}, but we use BN instead of LayerNorm~\cite{ba2016layer} after the conv layer as BN can be equivalently merged into the conv layer to eliminate its inference costs. We use another BN after the FFN, which can also be equivalently merged into the preceding layer (\ie, the second linear layer in FFN). The numbers of such blocks in the four stages are denoted by $\bm{N}$ ($N_1$,$N_2$,$N_3$,$N_4$), respectively. Following ConvNeXt-T, the vanilla architecture uses $C$=96 and $\bm{N}$=(3,3,9,3). By default, the last three stages use 13$\times$13 Dilated Reparam Block as the DW layer, which means $K$=13, $\bm{k}$=(5,7,3,3,3) and $\bm{r}$=(1,2,3,4,5); the first stage uses DW 3$\times$3 conv as the DW layer.

\noindent\textbf{Experimental settings and metrics}. It has been emphasized in the literature~\cite{ding2022scaling} that large-kernel ConvNets should be evaluated on downstream tasks, as their full potential may not be accurately reflected by the ImageNet accuracy alone. Therefore, except for the ImageNet-1K accuracy after 100-epoch training, we transfer the trained model with UPerNet~\cite{xiao2018unified} to ADE20K to examine its performance on semantic segmentation and report the single-scale mIoU after a 160k-iteration standard finetuning process~\cite{mmseg2020}. Besides the parameters and FLOPs, we test the actual throughput on an A100 GPU with a batch size of 128 and input resolution of 224$\times$224, which is measured in images per second (img/s). See the Appendix for detailed configurations.

\begin{figure}
		\begin{center}
			\includegraphics[width=1.0\linewidth]{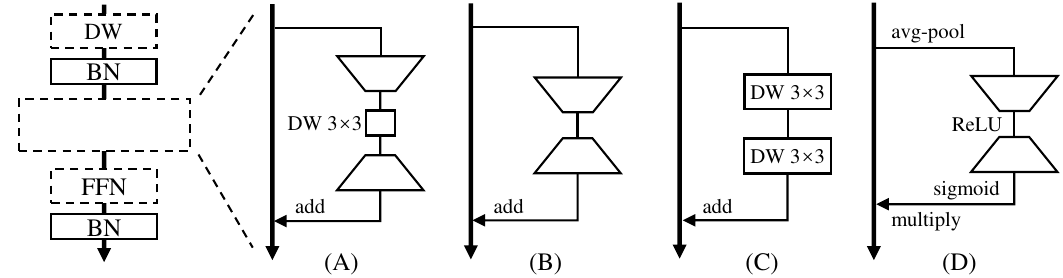}
			\vspace{-0.3in}
			\caption{Options of the extra structures to increase the depth.}
			\label{fig-se}
			\vspace{-0.25in}
		\end{center}
	\end{figure}

\noindent\textbf{Architectural Guideline 1 on Block Design: use efficient structures that perform both inter-channel communications and spatial aggregations to increase the depth}. We first seek to insert some structures to universally boost the model's representational capacity, which is required to comprise nonlinearity and efficient trainable transformations. We naturally try a bottleneck composed of a 1$\times$1 conv that reduces the channels to 1/4, a DW 3$\times$3 conv, and another 1$\times$1 conv to expand the channels back (Fig.~\ref{fig-se}). We use BN and ReLU after conv layers as a common practice. Table~\ref{table-guide1} shows that the performance improves with acceptable costs (+1.2 mIoU with 12\% slow down). The performance degrades as we remove the DW 3$\times$3 conv so that only two 1$\times$1 conv layers remain, or replace the bottleneck structure with two DW 3$\times$3 layers, suggesting that such structures require both spatial aggregation transformations and channel mixing. Motivated by this, considering that SE Block~\cite{hu2018squeeze} elegantly realizes both transformations in a more efficient way (\ie, global average pooling and nonlinear mapping of the pooled vectors), we try it also with 1/4 channel reduction and observe a better performance and higher throughput. We therefore use SE Block as a substructure of our block design in the following explorations.

	\begin{table}
		\caption{Models with different efficient extra structures to increase the depth. We report the ImageNet accuracy (Acc), ADE20K mIoU, and actual throughput (Img/s).}
		\label{table-guide1}
		\vspace{-0.3in}
		\begin{center}
        \resizebox{1.0\linewidth}{!}{
			\tiny
			\begin{tabular}{lccccc}
				\hline
			Extra structure	       & Params    &  FLOPs & Img/s  & Acc &	mIoU\\
				\hline
                None               & 31.3M   & 4.92G   &   1954    &   81.2    &   45.1 \\
                (A) Bottleneck          & 32.9M   & 5.18G   &   1716    &   81.5    &   46.3\\
                (B) Two 1$\times$1      & 32.9M   & 5.17G   &   1745    &   81.3    &   46.2\\
                (C) Two DW 3$\times$3   & 31.4M   & 4.96G   &   1659    &   81.3    &   45.4\\
                (D) SE Block            & 32.9M & 4.92G   &   1863      &   \textbf{81.6}    &   \textbf{46.5} \\
				\hline
			\end{tabular}
        }
		\end{center}
		\vspace{-0.3in}
	\end{table}

\noindent\textbf{Architectural Guideline 2 on Re-parameterization: use dilated small kernels to re-parameterize a large kernel}. For a fair comparison with Dilated Reparam Block, we try two variants with the same numbers of parallel branches composed of non-dilated layers with \textbf{A)} the same kernel sizes or \textbf{B)} the same equivalent kernel sizes. For our default setting of $K$=13, $\bm{r}$=(1,2,3,4,5), $\bm{k}$=(5,7,3,3,3), the kernel sizes of the five branches will be $\bm{k}$=(5,7,3,3,3) or (5,13,7,9,11) for the two variants, respectively. All the models end up with the same inference structure but the training structures differ. Table~\ref{table-reparam} shows lower performance of variants, suggesting that large kernel benefits from the parallel dilated conv layers' abilities to capture sparse patterns, rather than merely the extra small kernels (variant A) or the combination of different receptive fields (variant B). We use Dilated Reparam Block in the following explorations.~\footnote{While describing the architecture in this paper, using a $K$$\times$$K$ ($K$$\geq$9) conv means a $K$$\times$$K$ Dilated Reparam Block, unless otherwise noted.}

	\begin{table}
		\caption{Different forms of Structural Re-parameterization on the 13$\times$13 conv layers. We report the mean$\pm$std of three runs.}
		\label{table-reparam}
		\vspace{-0.3in}
		\begin{center}
        \resizebox{1.0\linewidth}{!}{
			\tiny
			\begin{tabular}{lccccc}
				\hline
			Re-param	            &   $\bm{k}$      &   $\bm{r}$   &      Acc &	mIoU\\
				\hline
                None                    &   N/A         &   N/A         &   81.44$\pm$0.04       &   45.78$\pm$0.05\\          
                Dilated Reparam  &   5,7,3,3,3 &  1,2,3,4,5            &   81.63$\pm$0.02    &   \textbf{46.37}$\pm$0.10 \\    
                Same kernel size    &   5,7,3,3,3 &  1,1,1,1,1          &   81.55$\pm$0.01    &   46.07$\pm$0.07\\             
                Same eq kernel size        &   5,13,7,9,11 &  1,1,1,1,1 &   81.59$\pm$0.02    &   46.17$\pm$0.04     \\        
				\hline
			\end{tabular}
   }
		\end{center}
		\vspace{-0.2in}
	\end{table}

\noindent\textbf{Architectural Guideline 3 on Kernel Size: decide kernel size according to the downstream task and usually use large kernels in middle- and high-level layers}. As introduced above, the baseline model uses 3$\times$3 conv in the first stage and 13$\times$13 in the last three stages. Table~\ref{table-guide3} shows that replacing the large kernels in the last three stages with 3$\times$3 or changing $K$ from 13 to 11 degrades the models, especially in the ADE20K mIoU, which highlights the significance of large kernels. Interestingly, using 13$\times$13 in Stage 1 or enlarging $K$ from 13 to 15 makes almost no difference in the ImageNet accuracy but reduces the ADE20K mIoU.

\noindent\textbf{Remark}. We argue that this phenomenon does not mean larger kernels result in lower feature quality. It is due to the structural priors of UPerNet, which takes the features extracted by the low-level layers of the backbone and assumes they should only encode local information so that combining them with the high-level features extracted from the last layers of the backbone results in better segmentation. With larger kernels in lower stages, the low-level features are no longer confined to small local areas so the UPerNet benefits less from combining them with the high-level features. We verify this explanation by making the UPerNet only use the high-level features (\ie, outputs of Stage 4) to evaluate the quality of the eventual features alone. Under this setting, $K$=15 delivers the best mIoU (42.7), the model with large kernels in Stage 1 performs as well as the baseline (42.4), and $K$=11 performs the worst (41.9). Such observations confirm that large kernels, even when they are used inappropriately, \emph{do not damage the feature quality} of ConvNet but \emph{merely make the low-level features less favorable for certain downstream models that require local low-level features}, suggesting we should decide the kernel size according to the specific downstream tasks and framework. In our specific use cases (\ie, representative image recognition tasks with common downstream frameworks), we employ 13$\times$13 kernels in the middle- and high-level stages by default.

	\begin{table}
		\caption{Models with different kernel sizes in the four stages denoted by S1 - S4. Numbers in parentheses are obtained with the UPerNet only taking the outputs of S4.}
		\label{table-guide3}
		\vspace{-0.3in}
		\begin{center}
        \resizebox{1.0\linewidth}{!}{
			\tiny
			\begin{tabular}{lcccccccc}
				\hline
			  S1  & S2   & S3  & S4  &  Params & FLOPs  & Img/s & Acc &  mIoU\\
				\hline
    3& 13 & 13 & 13 & 32.9M & 4.92G   &   1863    &   81.6    &   \textbf{46.5} (42.4)    \\ 
    \hline
                3& 11 & 11 & 11 &  32.6M   & 4.86G  &   1876      & 81.6      &   45.5 (41.9)    \\
                3&  3& 13 & 13    & 32.8M   &   4.85G   &   2006    & 81.7  &   46.1\\
                  3& 13 &  3& 13    & 32.4M   &   4.81G   &   2015    & 81.6  &   45.9\\
                  3& 13 & 13 & 3     & 32.5M   &   4.90G   &   1884    & 81.4  &   45.8\\                             
                 \hline
                 3& 15 & 15 & 15 &   33.3M  &  4.99G    &   1851    &    81.7   &   45.9 (\textbf{42.7})    \\
                 13 & 13 & 13 & 13 &    33.0M   &5.06G  &   1547    &   81.6    & 44.9 (42.4)  \\
                 \hline
                  
				\hline
			\end{tabular}
        }
		\end{center}
		\vspace{-0.3in}
	\end{table}

\noindent\textbf{Architectural Guideline 4 on the Scaling Rule: while scaling up the depth, the added blocks should use small kernels.} The scaling rule of existing large-kernel ConvNets follows the traditional ConvNets, \ie, stacking more large kernels to build up a deeper model, but we argue that a large-kernel ConvNet may not benefit from more large kernels. In this group of experiments (Table~\ref{table-guide4}), we scale up $N_3$ from 9 to 27, following ConvNeXt-S~\cite{liu2022convnet}. Considering that nine 13$\times$13 blocks may have already built up sufficient receptive field, we examine if the added blocks should also use large kernels. Specifically, we refer to the block with a Dilated Reparam Block as the \emph{Large Kernel Block (LarK Block)} and name a block that uses a DW 3$\times$3 conv as a \emph{Small Kernel Block (SmaK Block)} so that there are 3 SmaK Blocks in Stage 1 and 3/9/3 LarK Blocks in Stage 2/3/4 of the shallow model. While scaling up the depth of Stage 3, we tried the following options. \textbf{A)} All of the 27 blocks are LarK Blocks. \textbf{B)} We interleave SmaK with LarK Blocks so that Stage 3 has 14 LarK Blocks and 13 SmaK Blocks. \textbf{C)} We place two SmaK Blocks after a LarK Block so that the resultant model will have the same 9 LarK Blocks as before but 18 extra SmaK Blocks. \textbf{D)} We remove the DW 3$\times$3 layers in SmaK Blocks. Table~\ref{table-guide4} shows that scaling up the depth brings significant improvements, which is expected, and 9 LarK Blocks are sufficient. Though 27 LarK Blocks perform slightly better in the ADE20K mIoU, the inference speed is observably slowed down. Besides, the model without 3$\times$3 conv in SmaK Blocks shows significantly lower mIoU with only minor improvements in the throughput, suggesting such small kernels in SmaK Blocks are useful while scaling up the depth of large-kernel ConvNet as they increase the abstract hierarchy of spatial patterns, though they may not effectively enlarge the ERF~\cite{ding2022scaling,erf}. This observation supports our motivation to decouple the effects of conv layers in enlarging the ERF and extracting more complicated spatial patterns, as discussed in Sec.~\ref{sec:intro}. 

	\begin{table}
		\caption{Different numbers of LarK and SmaK Blocks in Stage 3.}
		\label{table-guide4}
		\vspace{-0.3in}
		\begin{center}
        \resizebox{1.0\linewidth}{!}{
			\tiny
			\begin{tabular}{lcccccccc}
				\hline
			  $N_3$   & LarK & SmaK     &  Params & FLOPs  & Img/s & Acc &  mIoU\\
				\hline
                9   &   9   &   0        & 32.9M & 4.92G   &   1863    &   81.6    &   \textbf{46.5}   \\
                \hline
                27  &   27  &   0         &   56.7M   &9.31G  &   1145   &   82.3   &   49.0\\
                27  &   14   &   13,  3$\times$3  &   55.9M   &9.15G  &   1229  &   82.3   &   48.8\\
                27  &   9   &   18,   3$\times$3  &   55.6M   &9.10G  &   1264   &   82.3   &   48.8\\
                \hline
                27  &   9   &   18, w/o 3$\times$3    &   55.5M   &9.08G  &   1289   &   82.2   &   47.8\\
				\hline
			\end{tabular}
        }
		\end{center}
		\vspace{-0.3in}
	\end{table}

\subsection{Architectural Specifications}\label{sec-}

Following our proposed guidelines, we instantiate a series of models (Table~\ref{table-instances}). For a fair comparison with ConvNeXt V2~\cite{woo2023convnext}, UniRepLKNet-A/F/P/N follows its configurations. We scale up the depth to build UniRepLKNet-T/S and scale up the width to construct UniRepLKNet-S/B/L/XL. 

	\begin{table}
		\caption{Architectural hyper-parameters of UniRepLKNet instances, including the number of blocks in the four stages $N_1, N_2, N_3, N_4$ and channels $C$ of the first stage. Stage 1 uses SmaK Blocks, and Stages 2 and 4 use LarK Blocks only. For Stage 3, \eg, ``9 + 18'' means 9 LarK Blocks and 18 SmaK Blocks.}
		\label{table-instances}
		\vspace{-0.3in}
		\begin{center}
        \resizebox{1.0\linewidth}{!}{
			\tiny
			\begin{tabular}{lcccccccc}
				\hline
			   &$N_1$ & $N_2$ & $N_3$ & $N_4$ & $C$ &  Params    \\
				\hline
                UniRepLKNet-A & 2   & 2     & 6 + 0     & 2     &   40   &  4.4M\\
                UniRepLKNet-F & 2   & 2     & 6 + 0     & 2     &   48   &  6.2M\\
                UniRepLKNet-P & 2   & 2     & 6 + 0     & 2     &   64   &  10.7M\\
                UniRepLKNet-N & 2   & 2     & 8 + 0     & 2     &   80   &  18.3M\\
                UniRepLKNet-T & 3   & 3     & 9 + 9     & 3     &   80   &  31.0M\\
                UniRepLKNet-S & 3   & 3     & 9 + 18    & 3     &   96   &  55.6M\\
                UniRepLKNet-B & 3   & 3     & 9 + 18    & 3     &   128   & 97.9M\\
                UniRepLKNet-L & 3   & 3     & 9 + 18    & 3     &   192   & 218.3M\\
                UniRepLKNet-XL & 3   & 3     & 9 + 18    & 3     &   256  & 386.4M\\
				\hline
			\end{tabular}
        }
		\end{center}
		\vspace{-0.3in}
	\end{table}

\subsection{Generalizing UniRepLKNet beyond Image}~\label{sec:mm-design}
To utilize the universal perception ability of UniRepLKNet, we preprocess the data of different modalities into $B\times C^\prime\times H\times W$ embedding maps, where $B$ is the batch size and $C^\prime$ is determined by the modality, and configure the input channel of the first layer of UniRepLKNet to $C^\prime$. For simplicity, the other parts of the models are the same as the UniRepLKNet initially designed for the image without any modality-specific customization. By doing so, we directly apply a ConvNet typically used for image tasks to deal with data of other modalities. In other words, the UniRepLKNet for image tasks can be seen as a general UniRepLKNet with $C^\prime$=3 and no such preprocessing. We introduce how to transform the data into such embedding maps as follows.

    \noindent\textbf{Time-series}. Let $L$ and $D$ be the length and dimensions of a time-series sequence $\boldsymbol{x}_T\in \mathbb{R}^{B\times L \times D}$, we adopt the embedding layer in Corrformer~\cite{wu2023interpretable} to split it into $n$ nodes then project it into a latent space $\mathbb{R}^{Bn \times L \times D^\prime}$ ($D^\prime$ and $n$ are configurable hyper-parameters of the embedding layer). Then we simply reshape it into a single-channel embedding map.
    \begin{equation}\label{eq:time:token}
        \begin{aligned}
             \boldsymbol{x}_T \in\mathbb{R}^{B\times L \times D } & \rightarrow \mathbb{R}^{Bn \times L \times \frac{D}{n}} \rightarrow \mathbb{R}^{Bn \times L \times D^\prime} \\ &\rightarrow \mathbb{R}^{Bn\times 1 \times H \times W }\, \text{s.t.} HW=LD^\prime.
        \end{aligned}
    \end{equation}
    \textbf{Audio}. Let $T$ and $F$ be the numbers of time frames and frequency bins, we use $\boldsymbol{x}_A \in \mathbb{R}^{B\times T\times F}$ to represent audio data. A sample is seen as a $1\times T\times F$ embedding map that resembles a single-channel image so $C^\prime$=1, $H$=$T$, $W$=$F$.
	\begin{equation}
		\boldsymbol{x}_A \in \mathbb{R}^{B\times T \times F} \rightarrow \mathbb{R}^{B\times 1 \times T \times F}.
		\label{eq:audio:token}
	\end{equation}
    \textbf{Point cloud}. Assume a sample comprises \(P\) points each represented by the X/Y/Z coordinates, we use a series of conv layers to generate three-view projections~\cite{zhang2023meta}. We configure the resolution of the generated projections to be 224 so that $H$=$W$=224, $C^\prime$=3.
 	\begin{equation}
		\boldsymbol{x}_P \in \mathbb{R}^{B\times P \times 3} \rightarrow  \mathbb{R}^{B\times 3 \times 224 \times 224 }\,.
		\label{eq:pcd:token}
	\end{equation}
    \textbf{Video}. We represent a video as $N_F$ frames and each frame is a $3\times h \times w$ image. We reshape it by merging the frame dimension into the height and width dimensions so that we obtain a representation that can be viewed as a single image created by laying out (\ie, concatenating) the $N_F$ frames. For example, in our experiments, we have $N_F$=16 and $h$=$w$=224 so that $H$=$W$=896. Generally,
    \begin{equation}
        \label{eq:video:token}
        \boldsymbol{x}_V\in\mathbb{R}^{B\times N_F\times3\times h \times w}\rightarrow \mathbb{R}^{B\times 3\times H\times W}\text{s.t.}\frac{HW}{hw}=N_F\,.
    \end{equation}

    \vspace{-0.2in}

\section{UniRepLKNet for Image Recognition}

\textbf{ImageNet classification}. Following ConvNeXt~\cite{liu2022convnet}, we use the widely adopted 300-epoch receipt to train UniRepLKNet-A/F/P/N/T/S on ImageNet-1K; we pretrain UniRepLKNet-S/B/L/XL on ImageNet-22K using the 90-epoch receipt and fine-tune with ImageNet-1K for 30 epochs (see the Appendix for details). As our goal is to develop models that \emph{run with high actual speed}, we evaluate the actual throughput on the same A100 GPU using a batch size of 128. Table~\ref{table-imgnet} shows the top-1 accuracy on the ImageNet-1K validation set where the results are sorted by the throughput. We split the results into seven segments for better readability. \textbf{1)} UniRepLKNet-A/F outperforms ConvNeXt-V2-A/F by 0.8/0.6 in the accuracy and runs 19\%/17\% faster, respectively. \textbf{2)} UniRepLKNet-P/N outperforms FastViT-T12/S12 and ConvNeXt V2-P/N by clear margins. \textbf{3)} UniRepLKNet-T outperforms multiple small-level competitors. \textbf{4)} UniRepLKNet-S outperforms a series of small-level and even base-level models in both speed and accuracy and runs almost as fast as InternImage-T. \textbf{5)} With ImageNet-22K pretraining, UniRepLKNet-S even approaches the accuracy of RepLKNet-31L and runs 3$\times$ as fast as the latter. UniRepLKNet-B outperforms CoAtNet-2 and DeiT III-B by clear margins. UniRepLKNet-L outperforms InternImage-L in both accuracy and throughput. \textbf{6)} On the XL-level, UniRepLKNet-XL outperforms in both accuracy and throughput, running more than 2$\times$ as fast as CoAtNet-3 and 3$\times$ as DeiT III-L.

\begin{table}[t!]
    \centering
    \renewcommand\arraystretch{0.89}
    \setlength{\tabcolsep}{0.9mm}
    \footnotesize
    \caption{\textbf{ImageNet classification}. Throughput is tested with an A100 GPU and batch size of 128. ``T/C'' denote transformer/ConvNet. ``$^\ddagger$" indicates ImageNet-22K~\cite{deng2009imagenet} pretraining.}
    \vspace{-0.1in}
    \begin{tabular}{l|c|c|c|c|c|c}
\hline
    \multirow{2}{*}{Method} & \multirow{2}{*}{Type} & Input & Params & FLOPs & Throughput & Acc \\
    & & size &(M)&(G)&(img/s)&(\%) \\ 
    \hline
    \rowcolor{gray!20}
    \textbf{UniRepLKNet-A}   &   C   &   $224^2$     &   4.4    &   0.6    &   \textbf{5942}    &   \textbf{77.0}    \\
    \rowcolor{gray!20}
    \textbf{UniRepLKNet-F}   &   C   &   $224^2$     &   6.2    &   0.9    &   \textbf{5173}    &    \textbf{78.6}       \\
    ConvNeXt V2-A~\cite{woo2023convnext}   &   C   &   $224^2$     &   3.7    &   0.5    &   5054    &   76.2        \\
    FastViT-T8~\cite{vasu2023fastvit}      &   T   &   $256^2$     &   3.6    &   0.7    &   5025    &   75.6    \\
    ConvNeXt V2-F~\cite{woo2023convnext}   &   C   &   $224^2$     &   5.2    &   0.8    &   4329    &   78.0        \\
    \hline
    \rowcolor{gray!20}
    \textbf{UniRepLKNet-P}   &   C   &   $224^2$   & 10.7 & 1.6    &   \textbf{3949}    &    \textbf{80.2}    \\
    FastViT-T12~\cite{vasu2023fastvit}      &   T   &   $256^2$    &   6.8     &   1.4    &   3407    &   79.1    \\
    ConvNeXt V2-P~\cite{woo2023convnext}   &   C   &   $224^2$     &   9.1    &   1.4   &   3339    &   79.7  \\
    FastViT-S12~\cite{vasu2023fastvit}      &   T   &   $256^2$    &   8.8     &   1.8    &   3162    &   79.8    \\
    \rowcolor{gray!20}
    \textbf{UniRepLKNet-N}   &   C   &   $224^2$ &  18.3&   2.8    &   \textbf{2807}    &    \textbf{81.6}\\
    ConvNeXt V2-N~\cite{woo2023convnext}   &   C   &   $224^2$     &   15.6   &   2.4   &   2405   &   81.2    \\
    \hline
    \rowcolor{gray!20}
    \textbf{UniRepLKNet-T}   &   C   &   $224^2$     &   31    &   4.9    &   \textbf{1804}   &   \textbf{83.2}    \\
    FastViT-SA24~\cite{vasu2023fastvit}      &   T   &   $256^2$   &   21    &   3.8    &   1670    &   82.6    \\
    PVTv2-B2~\cite{wang2022pvt} & T & $224^2$  & 25 & 4.0     &   1620    & 82.0 \\
    CoAtNet-0~\cite{dai2021coatnet} & T &$224^2$ & 25 & 4.2 &   1613& 81.6   \\
    DeiT III-S~\cite{touvron2022deit} & T &$224^2$  & 22 & 4.6  &   1485    &   81.4\\
    SwinV2-T/8~\cite{liu2022swin} & T &$256^2$ & 28 & 6   &   1406    & 81.8 \\
    SLaK-T~\cite{liu2022more} & C & $224^2$  & 30 & 5.0 & 1312 &82.5  \\
    InternImage-T~\cite{wang2023internimage}   & C     &$224^2$        & 30        & 5        &   1292   &    83.5 \\
    \hline
    \rowcolor{gray!20}
    \textbf{UniRepLKNet-S}   &   C   &   $224^2$     &   56    &   9.1    &   \textbf{1265}   &    \textbf{83.9}    \\
    ConvNeXt-S~\cite{liu2022convnet}  &   C   &   $224^2$     &   50  &   8.7 &   1182    &   83.1\\
    HorNet-T~\cite{rao2022hornet} & C & $224^2$ & 23 & 3.9 & 1162   &   83.0 \\
    FastViT-SA36~\cite{vasu2023fastvit}    &   T   &   $256^2$     &   30    &   5.6    &   1151    &   83.6     \\
    CoAtNet-1~\cite{dai2021coatnet} & T &$224^2$ & 42 & 8.4 &   969 & 83.3  \\
    SLaK-S~\cite{liu2022more} & C &$224^2$  & 55 & 9.8 & 967   &83.8 \\
    FastViT-MA36~\cite{vasu2023fastvit}    &   T   &   $256^2$     &   43    &   7.9    &   914     &   83.9    \\
    SwinV2-S/8~\cite{liu2022swin} & T &$256^2$ & 50 & 12  &   871     & 83.7\\
    RepLKNet-31B~\cite{ding2022scaling} & C &$224^2$  & 79 & 15.3 &    859 & 83.5\\
    PVTv2-B5~\cite{wang2022pvt} & T & $224^2$  & 82 & 11.8    &   802     & 83.8 \\
    
    \hline
    \hline

\rowcolor{gray!20}
    \textbf{UniRepLKNet-S}$^\ddagger$  & C & $384^2$   &    56 &  26.7    &   \textbf{435}         &   \textbf{86.4}  \\
    ConvNeXt-S$^\ddagger$~\cite{liu2022convnet}  &   C   &   $384^2$     &   50  &   25.5    & 415   &   85.8\\
\rowcolor{gray!20}
    \textbf{UniRepLKNet-B}$^\ddagger$  & C & $384^2$    &   98    &   47.2    &   \textbf{314}     &   \textbf{87.4}\\
    ConvNeXt-B$^\ddagger$~\cite{liu2022convnet}  & C &   $384^2$ &   89  &   45.1   & 304 &   86.8\\
    \hline
    \rowcolor{gray!20}
    \textbf{UniRepLKNet-L}$^\ddagger$  & C & $384^2$    &   218   &   105.4   &   \textbf{190}     &   \textbf{87.9}\\
    ConvNeXt-L$^\ddagger$~\cite{liu2022convnet} & C &$384^2$ & 198 & 101  &   185 & 87.5\\
    CoAtNet-2$^\ddagger$~\cite{dai2021coatnet} & T &$384^2$ & 75 & 49.8      &   163  & 87.1      \\
    RepLKNet-31L$^\ddagger$~\cite{ding2022scaling} & C &$384^2$ & 172   & 96.0    &  158 &   86.6\\
    InternImage-L$^\ddagger$~\cite{wang2023internimage}    & C &$384^2$ & 223 & 108 & 143    &   87.7 \\
    DeiT III-B$^\ddagger$~\cite{touvron2022deit} & T &$384^2$  & 87 & 55.5           & 138    & 86.7    \\
    \hline
    \rowcolor{gray!20}
    \textbf{UniRepLKNet-XL}$^\ddagger$  & C & $384^2$   &   386   &   187   &   \textbf{131}     &   \textbf{88.0}\\
    ConvNeXt-XL$^\ddagger$~\cite{liu2022convnet} & C &$384^2$ & 350 & 179 &   129 & 87.8\\
    
    HorNet-L$^\ddagger$~\cite{rao2022hornet} & C & $384^2$ & 202 & 102 & 127  &87.7 \\
    InternImage-XL$^\ddagger$~\cite{wang2023internimage}   & C &$384^2$ & 335 & 163 & 114    &   88.0 \\

    CoAtNet-3$^\ddagger$~\cite{dai2021coatnet} & T &$384^2$ & 168 & 107    &   103  & 87.6    \\
    SwinV2-L/24$^\ddagger$~\cite{liu2022swin} & T &$384^2$ & 197 & 115    &   88   & 87.6\\
    CoAtNet-4$^\ddagger$~\cite{dai2021coatnet} & T &$384^2$ & 275 & 190    &   58   & 87.9    \\
    DeiT III-L$^\ddagger$~\cite{touvron2022deit} & T &$384^2$  & 305 & 191         & 42     & 87.7    \\

    \hline
        
\end{tabular}
    \vspace{-0.2in}
    \label{table-imgnet}
\end{table}

\noindent\textbf{COCO object detection and instance segmentation}. We transfer the pretrained UniRepLKNets as the backbones of Cascade Mask R-CNN~\cite{he2017mask,cai2019cascade} and adopt the standard 3x (36-epoch) training configuration with MMDetection~\cite{mmdetection}. Table~\ref{tab:det} shows UniRepLKNet outperforms Swin, ConvNeXt, RepLKNet, and SLaK, which are representatives of ViTs, modern medium-kernel ConvNets, and existing large-kernel ConvNets, respectively, and shows comparable performance to InternImage~\cite{wang2023internimage}, which is a latest powerful architecture with deformable convolution.

\begin{table}[t]
        \centering
    \renewcommand\arraystretch{0.89}
    \setlength{\tabcolsep}{0.9mm}
    \footnotesize
    \caption{\textbf{Object detection on COCO validation set}. FLOPs are measured with 1280$\times$800 inputs. ``$^\ddagger$" ImageNet-22K pretraining.}
    \vspace{-0.1in}
    
\begin{tabular}{l|c|c|c|c}
\hline
    Method & Params (M) &  FLOPs (G)  & $\text{AP}^{\text{box}}$  &  $\text{AP}^{\text{mask}}$   \\
    \hline
    \rowcolor{gray!20}
    \textbf{UniRepLKNet-T}   &          89  &   749     &   \textbf{51.8}    &   \textbf{44.9}     \\
    Swin-T~\cite{liu2021swin}      &   86   &   745     &   50.4    &   43.7\\
    ConvNeXt-T~\cite{liu2022convnet}  &   86   &   741      &   50.4    &   43.7\\
    SLaK-T~\cite{liu2022more}      &   -   &   -        &   51.3    &   44.3\\
    
    \hline
    \rowcolor{gray!20}
    \textbf{UniRepLKNet-S}       &   113  &   835    &   \textbf{53.0}      &     \textbf{45.9}      \\ 
    Swin-S~\cite{liu2021swin}      &   107   &   838    &   51.9    &   45.0\\
    ConvNeXt-S~\cite{liu2022convnet}  &   108   &   827     &   51.9    &   45.0\\
    
    \hline
    \rowcolor{gray!20}
    \textbf{UniRepLKNet-S}$^\ddagger$      &   113  &   835   &   \textbf{54.3}      &     \textbf{47.1}   \\
    \rowcolor{gray!20}
    \textbf{UniRepLKNet-B}$^\ddagger$      &   155  &   978   &    \textbf{54.8}    &   \textbf{47.4}   \\
    Swin-B$^\ddagger$~\cite{liu2021swin}      &   145   &   982     &   53.0    &   45.8    \\
    ConvNeXt-B$^\ddagger$~\cite{liu2022convnet}  &   146   &   964     &   54.0    &   46.9    \\
    RepLKNet-31B$^\ddagger$~\cite{ding2022scaling} &   137 &   965 &52.2   &   45.2    \\
    \hline
    \rowcolor{gray!20}
    \textbf{UniRepLKNet-L}$^\ddagger$      &   276  &   1385  & {55.8}   &   {48.4}      \\
    Swin-L$^\ddagger$~\cite{liu2021swin}  &   253 &   1382    &   53.9    &   46.7    \\
    ConvNeXt-L$^\ddagger$~\cite{liu2022convnet}  &  255 &1354   &   54.8    &   47.6    \\
    RepLKNet-31L$^\ddagger$~\cite{ding2022scaling}    &   229 &1321   &   53.9    &   46.5    \\
    InternImage-L$^\ddagger$~\cite{wang2023internimage}   &   277   &   1399     &\textbf{56.1}   &   \textbf{48.5}   \\
    \hline
    \rowcolor{gray!20}
    \textbf{UniRepLKNet-XL}$^\ddagger$  &   443  &   1952   &   \textbf{56.4}    &   \textbf{49.0}   \\
    InternImage-XL$^\ddagger$~\cite{wang2023internimage}  &   387    &   1782  & 56.2 &   48.8  \\
    ConvNeXt-XL$^\ddagger$~\cite{liu2022convnet} &   407   &   1898    &   55.2    &   47.7     \\
    \hline
\end{tabular}
    \label{tab:det}
    \vspace{-0.1in}
\end{table}

\noindent\textbf{ADE20K semantic segmentation}. We use the pretrained UniRepLKNets as the backbones of UPerNet~\cite{xiao2018unified} on ADE20K~\cite{zhou2019semantic} and adopt the standard 160k-iteration training receipt with MMSegmentation~\cite{mmseg2020}. Table~\ref{tab:seg} reports the mIoU on the validation set. Impressively, UniRepLKNet outperforms InternImage and the other models.

\begin{table}[t]
    \centering
    \setlength{\tabcolsep}{1.3mm}
    \footnotesize
    \caption{\textbf{Semantic segmentation on ADE20K validation set}. The FLOPs are measured with 512$\times$2048 or 640$\times$2560 inputs according to the crop size. ``SS'' and ``MS" mean single- and multi-scale testing, respectively. ``$^\ddagger$" ImageNet-22K~\cite{deng2009imagenet} pretraining.}
    \vspace{-0.1in}
    \begin{tabular}{l|c|c|c|cc}
    \hline
    \multirow{2}{*}{Method} & Crop & Params & FLOPs & mIoU & mIoU\\
    	& size & (M)&(G) & (SS) & (MS)   \\
    \hline
    \rowcolor{gray!20}
    \textbf{UniRepLKNet-T}    &   512$^2$    & 61   &   946   &   \textbf{48.6}    &   \textbf{49.1}   \\ 
    	Swin-T~\cite{liu2021swin} & 512$^2$ & 60  & 945 & 44.5 & 45.8 \\
    	ConvNeXt-T~\cite{liu2022convnet} & 512$^2$ & 60 & 939 & 46.0 & 46.7 \\
    	SLaK-T~\cite{liu2022more} & 512$^2$ & 65 & 936 & 47.6 & - \\
    	InternImage-T~\cite{wang2023internimage} & 512$^2$ & 59 & 944 & 47.9 & 48.1  \\
     \rowcolor{gray!20}
     
    	\hline
     \rowcolor{gray!20}
        \textbf{UniRepLKNet-S}    &   512$^2$   &   86   &   1036   &   \textbf{50.5}    &   \textbf{51.0}   \\
        Swin-S~\cite{liu2021swin} & 512$^2$ & 81 &  1038 &  47.6 &  49.5 \\
        ConvNeXt-S~\cite{liu2022convnet}  & 512$^2$ &82 & 1027  & 48.7 & 49.6  \\
        SLaK-S~\cite{liu2022more} & 512$^2$ &91 & 1028 & 49.4 & - \\
        InternImage-S~\cite{wang2023internimage} & 512$^2$ & 80 & 1017 & 50.1 & 50.9 \\

        \hline
        \rowcolor{gray!20}
        \textbf{UniRepLKNet-S}$^\ddagger$    &   640$^2$    &  86 & 1618  &   \textbf{51.9}  &   \textbf{52.7}\\      
        \rowcolor{gray!20}
        \textbf{UniRepLKNet-B}$^\ddagger$    &   640$^2$    &  130  & 1850 &   \textbf{53.5}    &   \textbf{53.9}\\
        Swin-B$^\ddagger$~\cite{liu2021swin}    &   640$^2$ &   121 &   1841   & 50.0   &   51.7\\
        ConvNeXt-B$^\ddagger$~\cite{liu2022convnet}      &   640$^2$     &   122     &   1828    &    52.6   &   53.1\\
        RepLKNet-31B$^\ddagger$~\cite{ding2022scaling}   &   640$^2$       &   112 &   1829    &   51.5    &   52.3 \\ 
        \hline
        \rowcolor{gray!20}
        \textbf{UniRepLKNet-L}$^\ddagger$    &   640$^2$    &   254  &   2507   &   \textbf{54.5}    &   \textbf{55.1}   \\
        Swin-L$^\ddagger$~\cite{liu2021swin} 
        & 640$^2$ & 234 & 2468 & 52.1 & 53.5 \\
        RepLKNet-31L$^\ddagger$~\cite{ding2022scaling} 
        & 640$^2$ & 207 & 2404 & 52.4 & 52.7 \\
        ConvNeXt-L$^\ddagger$~\cite{liu2022convnet} 
        & 640$^2$ & 235 & 2458 & 53.2 & 53.7 \\

        InternImage-L$^\ddagger$~\cite{wang2023internimage}
        & 640$^2$ & 256  & 2526  & 53.9 & 54.1 \\
        \hline
        \rowcolor{gray!20}
        \textbf{UniRepLKNet-XL}$^\ddagger$    &   640$^2$    &   425  &   3420   &   \textbf{55.2}    &   \textbf{55.6}   \\
        ConvNeXt-XL$^\ddagger$~\cite{liu2022convnet}
        & 640$^2$ & 391 & 3335 & 53.6 & 54.0 \\
        InternImage-XL$^\ddagger$~\cite{wang2023internimage}
        & 640$^2$ & 368 & 3142 & 55.0 & 55.3 \\

        \hline
               
    \end{tabular}
    \label{tab:seg}
    \vspace{-0.2in}
\end{table}

\section{Universal Perception on other Modalities}

\begin{table}
\centering
	
\caption{\textbf{Time-series forecasting} performance on Global Temperature and Wind Speed Forecasting challenge. UniRepLKNet delivers a new state-of-the-art performance in Mean Squared Error (MSE) and Mean Absolute Error (MAE). GFS~(\url{https://www.ncei.noaa.gov/}) stands for the Global Forecasting System.}
\label{tab:time}
\vspace{-0.1in}
\resizebox{1\linewidth}{!}{
\begin{tabular}{lcccccc}
	\hline
	\multirow{2}{*}{Method}            & \multirow{2}{*}{Type}   & \multirow{2}{*}{Params}              & \multicolumn{2}{c}{Temperature}      & \multicolumn{2}{c}{Wind speed}      \\
        \cline{4-7}
        & & &  $\text{MSE}\downarrow$ & $\text{MAE}\downarrow$
        & $\text{MSE}\downarrow$ & $\text{MAE}\downarrow$ \\
	\hline
        \multicolumn{3}{@{\;}l}{\bf Statistics-based}\\
        Holt–Winters~\cite{hyndman2017forecasting}   & - & - 
        & 13.241 &2.262 &5.912 & 1.664 \\
        Prophet~\cite{taylor2018forecasting} & - & - 
        & 11.626 & 2.946 & 9.691 & 2.382 \\
        GDBT~\pub{NeurIPS'17}~\cite{ke2017lightgbm}  & - & -  
        & 9.706 & 2.214 & 4.101 & 1.417 \\
        \hline
        \multicolumn{3}{@{\;}l}{\bf Numerical Simulation}\\
	GFS (reanalysis) & - & -  
        & 14.933 & 2.287 & 9.993 & 2.340 \\
        ERA5 (reanalysis)~\cite{hersbach2020era5} & - & -   
        & 13.448 & 1.908 & 4.999 & 1.587 \\
        DeepAR~\cite{salinas2020deepar} & - & - 
        & 32.249 & 4.262 & 5.248 & 1.602 \\
        N-BEATS~\cite{oreshkin2019n} & - & - 
        & 9.203 & 2.117 & 4.124 & 1.390 \\
	\hline
        \multicolumn{3}{@{\;}l}{\bf Deep Learning Specialist}\\
        StemGNN~\pub{NeurIPS'20}~\cite{cao2020spectral} 
        & GNN & 180M & 13.926 & 2.746 & 4.066 & 1.389 \\
        Pyraformer~\pub{ICLR'21}~\cite{liu2021pyraformer} 
        & Transformer  & 158M & 23.326 & 3.669 & 4.614 & 1.514  \\
        Corrformer~\pub{Nat. Mach. Intell.'23}~\cite{wu2023interpretable} 
        & Transformer & 155M & 7.709 & 1.888 & 3.889 & 1.304 \\
        \hline
        \multicolumn{3}{@{\;}l}{\bf Generalist}\\
        \rowcolor{gray!20}
	UniRepLKNet-S & ConvNet & {132M} & \textbf{7.602} & \textbf{1.832} & \textbf{3.865} & \textbf{1.301} \\
	\hline
\end{tabular}
}
\vspace{-0.1in}
\end{table}

\textbf{Time-series}. Following Corrformer~\cite{wu2023interpretable}, we conduct experiments on the Global Temperature and Wind Speed Forecasting challenge~\footnote{\url{https://codeocean.com/capsule/0341365/tree/v1}} using the dataset collected from the National Centers for Environmental Information (NCEI). This huge-scale dataset contains hourly averaged wind speed and temperature data from 3,850 stations with different geographical scales and densities, spanning from 2019 to 2021. For a fair comparison with Corrformer, which was the previous state-of-the-art method, we use its embedding layer (as introduced in Sec.~\ref{sec:mm-design}) and decoder and only replace its encoder transformer with UniRepLKNet-S. We also compare UniRepLKNet-S against a wide range of methods, including statistical and numerical approaches. Table~\ref{tab:time} shows UniRepLKNet delivers a new state-of-the-art forecasting precision, achieving the lowest errors of 7.602, 1.832, 3.865, and 1.301 for MSE and MAE in forecasting global temperature and wind speed, respectively, with fewer parameters than existing deep learning methods. It is particularly noteworthy that UniRepLKNet, a generalist model, outperforms time-series specialists such as Pyraformer~\cite{liu2021pyraformer} and Corrformer~\cite{wu2023interpretable} in both precision and efficiency. The significant advantages of UniRepLKNet open up new avenues for architectural discussions in time-series forecasting, presenting a viable alternative to transformer models.

\noindent{\textbf{Audio}}. We use Speech Commands V2~\cite{warden2018speech}, which contains 105,829 one-second recordings of 35 common speech commands. Table~\ref{tab:audio} shows UniRepLKNet seamlessly adapts to audio and delivers an impressive accuracy of 98.5\%, even without pretraining. Compared to transformers such as AST~\cite{gong2021ast} and Audio-MAE~\cite{huang2022masked}, UniRepLKNet stands out with fewer parameters. Compared to previous ConvNets designed for audio, UniRepLKNet achieves better performance without customizations to the structure, highlighting the untapped potential of ConvNets in the realm of audio.

\begin{table}[t]
    \caption{\textbf{Audio recognition} on Speech Commands V2 dataset.}
    \vspace{-0.1in}
    \label{tab:audio}
    \centering
    \resizebox{0.98\linewidth}{!}{
    \begin{tabular}{lcccc}
        \hline
        Method 	& Pretrain & Type &  Acc. (\%) 	& {Params}\\
        \hline
        PANNS~\cite{kong2020panns} & - & ConvNet & 61.8   & -\\
        PSLA~\cite{gong2021psla} & IN-1K & ConvNet & 96.3 &  -\\
        AST~\cite{gong2021ast} &AS-2M & Transformer & 96.2  & 86.9M \\
        SSAST ~\cite{gong2022ssast}  &AS-2M & Transformer & 97.8  & 89.3M \\
        Audio-MAE~\cite{huang2022masked} &AS-2M & Transformer & 98.3  & 86.2M \\
        Meta-Transformer~\cite{zhang2023meta} & LAION-2B & Transformer & 97.0  & 86.6M\\
        \hline
        \rowcolor{gray!20}
        $\text{UniRepLKNet-S}$ & - & ConvNet & \textbf{98.5}  & \textbf{55.5M} \\
        \hline
    \end{tabular}
    }
    \vspace{-0.2in}
\end{table}

\noindent{\textbf{Video}}. Kinetics-400~\citep{kay2017kinetics} contains 240k training videos and 20k validation videos, spanning 400 classes for action recognition. Though the top-1 accuracy of 54.8\% is somewhat behind state-of-the-art architectures like MViT~\cite{li2022mvitv2}, we note that UniRepLKNet is a generalist model without pretraining. Compared to the latest generalist methods, ImageBind~\cite{girdhar2023imagebind} and Meta-Transformer~\cite{zhang2023meta}, UniRepLKNet shows higher accuracy and requires no pretraining. 

\begin{table}[t]
    \centering
    \caption{\textbf{Video recognition} accuracy on Kinetics-400.}
    \vspace{-0.1in}
    \label{tab:video}
    \centering
    \resizebox{0.93\linewidth}{!}{
        \begin{tabular}{lcccc}
            \hline
            Method  &  Pretrain & Type &  Acc (\%) & Params \\ \hline
            \multicolumn{3}{@{\;}l}{\bf Specialist}\\
            SlowFast-101~\cite{feichtenhofer2019slowfast} & IN-1K & ConvNet+RNN & 79.8 & 62.8M \\
            MViTv2-B~\cite{li2022mvitv2} & IN-1K & Transformer & 81.2 & 51.2M  \\
            TimeSFormer~\cite{bertasius2021space} & K400 & Transformer & 80.7 & 122M\\
            \hline
            \multicolumn{3}{@{\;}l}{\bf Generalist}\\
            Meta-Transformer~\cite{zhang2023meta} & LAINON-2B & Transformer & 47.3 & 86.9M\\
            ImageBind~\cite{girdhar2023imagebind} & CLIP Data& Transformer & 50.0 & 632M\\
            \hline
            \rowcolor{gray!20}
            UniRepLKNet-S & - & ConvNet & 54.8 & 55.5M \\
            \hline
            \end{tabular}
        }
\vspace{-0.1in}
\end{table}

\begin{table}[t]
\centering
\caption{\textbf{Point cloud analysis} on ModelNet-40 dataset.}
\label{tab:pcd}
\vspace{-0.1in}
\centering
\resizebox{0.78\linewidth}{!}{
\small
\begin{tabular}{lcccc}
	\hline
	\multirow{2}{*}{Method}            & \multirow{2}{*}{Type}          & \multicolumn{2}{c}{ModelNet-40}   \\
        &   & mAcc (\%)      & OA (\%)    \\
	\hline
	PointNet~\cite{qi2017pointnet}  & MLP     & 86.0          & 89.2          \\
	PointNet++~\cite{qi2017pointnet++}  & MLP & -             & 91.9        \\
        \hline
	PointConv~\cite{wu2019pointconv}  & ConvNet     &-             &92.5         \\
	KPConv~\cite{thomas2019kpconv}    & ConvNet    &-             &92.9  \\
	DGCNN~\cite{wang2019dynamic}     & ConvNet    & 90.2         & 92.9   \\
	\hline
        OpenShape~\cite{liu2023openshape} & Transformer & 83.4 & -  \\
 \rowcolor{gray!20}
	UniRepLKNet-S & ConvNet & \textbf{90.3} & \textbf{93.2}  \\
	\hline
\end{tabular}
}
\vspace{-0.1in}
\end{table}

\noindent{\textbf{Point cloud}}. We explore the versatility of UniRepLKNet by assessing its proficiency in learning 3D patterns, extending beyond the conventional 2D signals of images and audio. We use the ModelNet-40~\cite{wu2015modelnet} 3D shape classification task with 9,843/2,468 training/validation samples of CAD models from 40 classes. Table~\ref{tab:pcd} shows UniRepLKNet achieves an Overall Accuracy (OA) of 93.2\% and a mean Accuracy (mAcc) of 90.3\%, surpassing existing ConvNet-based models specialized for point cloud. Such outcomes highlight the potential of further developing ConvNets in this domain.

\noindent{\textbf{Impact of kernel size on the performance}}. To investigate the influence of different kernel sizes on performance, we compare UniRepLKNet with models of smaller kernels. We adopted the same modality-specific preprocessing approaches and training configurations for a fair comparison. We take ResNet-101 as a representative small-kernel ConvNet because it has comparable parameters to UniRepLKNet-S. Table~\ref{tab:mm_ablation} shows large kernels are crucial for universal perception, at least in our specific cases. 
\begin{table}[t]
    \centering
    \caption{Universal perception performance with other ConvNets or UniRepLKNet with a smaller kernel size.}
    \vspace{-0.1in}
    \centering
    \resizebox{0.93\linewidth}{!}{
    \begin{tabular}{lcccc}
         \toprule
         \multirow{2}{*}{Modality} &  Time-Series & Point Cloud & Audio & Video \\
         \cline{2-5} & MAE$\downarrow$ & OA (\%) & Acc (\%) & Acc (\%)\\
         \hline
         ResNet-101~\cite{he2016deep} (K=3)    & 7.846 & 92.6 &  73.6 & 41.3 \\ \hline
         ConvNeXt-S~\cite{liu2022convnet} (K=7) &  7.641 & 92.7 & 94.3 & 48.5\\  \hline
         UniRepLKNet-S (K=11) &  7.751 & 92.9 & 94.7   & 51.7\\
         \rowcolor{gray!20}
         UniRepLKNet-S (K=13) & \textbf{7.602} & \textbf{93.2} & \textbf{98.5} & \textbf{54.8}  \\
         \bottomrule
    \end{tabular}
    }
    \label{tab:mm_ablation}
    \vspace{-0.2in}
\end{table}

\section{Conclusion}
UniRepLKNet shows a leading performance in image recognition and achieves remarkable results even on modalities such as audio and time-series data, outperforming multiple specialist models on those modalities. Such results signify a \emph{``\textbf{comeback}''} for ConvNet in its original domain and showcase large-kernel ConvNet's potential to \emph{``\textbf{conquer}''} new territories. The limitations are noticeable, \eg, the dilated branches require more training resources, which may be upgraded with simpler~\cite{cai2023refconv} or gradient~\cite{ding2022re} re-parameterization; the applications to large vision-language models~\cite{clip,liu2024visual,wang2023makes}, cross-attention-based scenarios~\cite{zhang2023online,chen2021crossvit}, and generation tasks~\cite{podell2023sdxl,zhang2024interactivevideo} remain under-explored.

\noindent\textbf{Acknowledgements.}
This work is partially supported by the National Natural
Science Foundation of China (Grant No. 8326014).

{
    \small
    \bibliographystyle{ieeenat_fullname}
    \bibliography{main}
}

\newpage

\clearpage
\setcounter{page}{1}
\maketitlesupplementary

\section*{Appendix A: General Transformation from Dialted Convolution to Non-dilated Large-Kernel Convolution}

Since \emph{ignoring pixels of the input is equivalent to inserting extra zero entries into the conv kernel}, \emph{a dilated conv layer with a small kernel can be equivalently converted into a non-dilated layer with a sparse larger kernel}. Let $k$ be the kernel size and $r$ be the dilation rate of the dilated layer, by inserting zero entries, the kernel size of the corresponding non-dilated layer will be $(k-1)r+1$, which is referred to as the \emph{equivalent kernel size} for brevity. 

As discussed in the paper, to eliminate the inference costs of the extra dilated conv layers in the Dilated Reparam Block, we propose to equivalently transform the whole block into a single non-dilated conv layer for inference. As discussed before, let $k$ and $r$ be the kernel size and dilation rate, respectively, the transformation from a dilated conv layer's kernel $\mathrm{W}\in\mathcal{R}^{k\times k}$ to a non-dilated layer's kernel $\mathrm{W}^\prime\in\mathcal{R}^{((k-1)r+1)\times ((k-1)r+1)}$ can be elegantly realized by a transpose convolution with a stride of $r$ and an identity kernel $\mathrm{I}\in\mathcal{R}^{1\times1}$, which is scalar 1 but viewed as a kernel tensor. That is
\begin{equation}\label{eq-merge_supp}
    \mathrm{W}^\prime = \mathtt{conv\_transpose2d}(\mathrm{W}, \mathrm{I}, \text{stride}=r) \,.
\end{equation}

In general cases with multi-channel conv layers, let the input channels, output channels, and number of groups be $c_{\text{in}}$, $c_{\text{out}}$, and $g$, respectively, we denote the kernel by a 4D tensor whose shape is $c_{\text{out}} \times \frac{c_{\text{in}}}{g} \times k\times k$. 

\textbf{1)} For a multi-channel depthwise (DW) layer, the transformation is easily generalized from 2D to 4D - the identity kernel $\mathrm{I}$ is viewed as a 4D tensor $\mathrm{I}\in\mathcal{R}^{1\times1\times1\times1}$ and we still follow function~\ref{eq-merge_supp} to derive the equivalent kernel by transpose convolution.

\textbf{2)} For non-DW cases (\ie, $g < c_{\text{in}}$), the transformation can be seen as splitting the kernel into slices (which can each be seen as a DW kernel), converting the slices respectively, and concatenating the resultant non-dilated slices up. We present the code in pytorch (Fig.~\ref{fig:code}) and a test case demonstrating the equivalency (Fig.~\ref{fig:testcase}).

\begin{figure*}
    \begin{lstlisting}[language=Python]
import torch
import torch.nn as nn
import torch.nn.functional as F

def convert_dilated_to_nondilated(kernel, dilate_rate):
    identity_kernel = torch.ones((1, 1, 1, 1))
    if kernel.size(1) == 1:
        #   This is a DW kernel
        dilated = F.conv_transpose2d(kernel, identity_kernel, stride=dilate_rate)
        return dilated
    else:
        #   This is a dense or group-wise (but not DW) kernel
        slices = []
        for i in range(kernel.size(1)):
            dilated = F.conv_transpose2d(kernel[:,i:i+1,:,:], identity_kernel, stride=dilate_rate)
            slices.append(dilated)
        return torch.cat(slices, dim=1)
    \end{lstlisting}
    \caption{Pytorch code to convert a dilated conv layer's small kernel to a non-dilated layer's larger sparse kernel.}
    \label{fig:code}
\end{figure*}

\begin{figure*}
    \begin{lstlisting}[language=Python]
def test_equivalency(in_channels, out_channels, groups, large_kernel_size, small_conv_r, small_conv_k):
    equivalent_kernel_size = small_conv_r * (small_conv_k - 1) + 1
    large_conv = nn.Conv2d(in_channels, out_channels, kernel_size=large_kernel_size,
                           padding=large_kernel_size // 2, groups=groups, bias=False)
    dilated_conv = nn.Conv2d(in_channels, out_channels, kernel_size=small_conv_k, 
    padding=equivalent_kernel_size // 2,
                             dilation=small_conv_r, groups=groups, bias=False)
    H, W = 19, 19
    x = torch.rand(2, in_channels, H, W)
    origin_y = large_conv(x) + dilated_conv(x)
    equivalent_kernel = convert_dilated_to_nondilated(dilated_conv.weight.data, small_conv_r)
    rows_to_pad = large_kernel_size // 2 - equivalent_kernel_size // 2
    merged_kernel = large_conv.weight.data + F.pad(equivalent_kernel, [rows_to_pad] * 4)
    equivalent_y = F.conv2d(x, merged_kernel, bias=None, padding=large_kernel_size // 2, groups=groups)
    print('relative error:', (equivalent_y - origin_y).abs().sum() / origin_y.abs().sum())

test_equivalency(in_channels=4, out_channels=4, groups=1, 
    large_kernel_size=13, small_conv_r=3, small_conv_k=3)
    \end{lstlisting}
    \caption{A test case demonstrating the equivalency of the transformation.}
    \label{fig:testcase}
\end{figure*}

\section*{Appendix B: Training Configurations}

We present the detailed training configurations for image classification, object detection, and semantic segmentation. We have publicly released a reproducible training script and trained weights for every model on GitHub.

\noindent\textbf{ImageNet image classification.} The training configurations for the ImageNet-1K-only results shown in Section 4 are presented in Table \ref{tab:supp_cls_1k}. These configurations are similar to common practices. For the experiments in Section 3, we use the same configurations, except that the training epochs are set to 100 and the drop path rate is set to 0.1. For the models pretrained with ImageNet-22K and then finetuned on ImageNet-22K, the configurations are shown in Table \ref{tab:supp_cls_1k}. Note that we follow the configurations adopted by ConvNeXt for a fair comparison with ConvNeXt-S/B, and the configurations used by InternImage for a fair comparison with InternImage-L/XL (the results with ImageNet-22K-pretrained InternImage-S/B were not reported).

\begin{table*}[t]
    \centering
    \renewcommand\arraystretch{1.0}
    \footnotesize
    \caption{\textbf{Detailed training configurations of ImageNet-1K-only models.} Apart from the configurations shown in the table, we use random left-right flipping, random resized crop, color jitter of 0.4, Auto-augment, and no repeated augmentation for every model.}
    \resizebox{0.98\linewidth}{!}{
\begin{tabular}{@{\ }l|c|c|c|c|c|c}
\hline
settings & UniRepLKNet-A & UniRepLKNet-F & UniRepLKNet-P & UniRepLKNet-N & UniRepLKNet-T & UniRepLKNet-S \\
\hline
input scale & 
224 & 
224 &
224 & 
224 & 
224 &
224 \\
batch size & 
4096 & 
4096 &
4096 & 
4096 & 
4096 &
4096 \\
optimizer &
AdamW & 
AdamW &
AdamW &
AdamW &
AdamW &
AdamW \\
LR      & 
4$\times10^{-3}$ & 
4$\times10^{-3}$ &
4$\times10^{-3}$ & 
4$\times10^{-3}$ & 
4$\times10^{-3}$ &
4$\times10^{-3}$ \\
LR schedule& 
cosine  &
cosine & 
cosine & 
cosine & 
cosine &
cosine \\
weight decay     &
0.05  & 
0.05  & 
0.05 & 
0.05 &
0.05 &
0.05 \\
warmup epochs & 
5 &
5 &
5 & 
5 &
5 &
5 \\
epochs & 
300 &
300 &
300 & 
300 &
300 &
300  \\
\hline
mixup alpha  & 
0.3 & 
0.3 & 
0.3 &
0.5 & 
0.8 &
0.8 \\
cutmix alpha &
0.3 & 
0.3 & 
0.3 &
0.5 & 
1.0 &
1.0 \\
erasing prob. &
0.25    &
0.25   &
0.25 &
0.25 &
0.25 & 
0.25 \\

\hline
label smoothing $\varepsilon$ & 
0.1 & 
0.1 &
0.1  &
0.1 & 
0.1  &
0.1  \\
drop path rate & 
0.0 & 
0.0 & 
0.1 & 
0.1 &
0.2 &
0.4 \\
\hline
\end{tabular}
}
    \label{tab:supp_cls_1k}
\end{table*}

\begin{table*}[t]
    \centering
    \renewcommand\arraystretch{1.0}
    \footnotesize
        \caption{\textbf{Detailed training configurations of models pretrained with ImageNet-22K (IN-22K pt) and then finetuned on ImageNet-1K (IN-1K ft).} Apart from the configurations shown in the table, we use random left-right flipping, random resized crop, color jitter of 0.4, Auto-augment, and no repeated augmentation for every model.}
    
\resizebox{0.98\linewidth}{!}{
\begin{tabular}{@{\ }l|cc|cc|cc|cc}
\hline
\multirow{2}{*}{settings} & \multicolumn{2}{c|}{UniRepLKNet-S} & \multicolumn{2}{c|}{UniRepLKNet-B} & \multicolumn{2}{c|}{UniRepLKNet-L} & \multicolumn{2}{c}{UniRepLKNet-XL} \\
\cline{2-9}
& 
IN-22K pt & 
IN-1K ft &
IN-22K pt & 
IN-1K ft &
IN-22K pt & 
IN-1K ft &
IN-22K pt & 
IN-1K ft \\
\hline
input scale & 
224 & 
384 &
224 & 
384 & 
192 & 
384 &
192 & 
384 \\
batch size & 
4096 &
512 &
4096 &
512 &
4096 &
512 &
4096 &
512 \\
optimizer &
AdamW & 
AdamW &
AdamW &
AdamW &
AdamW &
AdamW & AdamW & AdamW\\
LR      & 
4$\times10^{-3}$ &
5$\times10^{-5}$ &
4$\times10^{-3}$ &
5$\times10^{-5}$ & 
4$\times10^{-3}$ &
5$\times10^{-5}$ & 
4$\times10^{-3}$ &
5$\times10^{-5}$ \\
LR schedule& 
cosine  &
cosine & 
cosine & 
cosine & 
cosine &
cosine & cosine & cosine\\
weight decay     &
0.05  &
1$\times10^{-8}$ & 
0.05  &
1$\times10^{-8}$ & 
0.05  &
1$\times10^{-8}$ & 
0.05  &
1$\times10^{-8}$ \\ 
warmup epochs & 
5 &
0 &
5 &
0 &
5 &
0 &
5 &
0 \\
epochs & 
90 &
30 &
90 &
30 &
90 &
20 &
90 &
20 \\
\hline

mixup alpha  & 
0.8 & 
0.0 &
0.8 & 
0.0 &
0.8 & 
0.0 &
0.8 & 
0.0 \\
cutmix alpha &
1.0 & 
0.0 &
1.0 & 
0.0 &
1.0 & 
0.0 &
1.0 & 
0.0 \\
erasing prob. &
0.25    &
0.25   &
0.25 &
0.25 &
0.25    &
0.25   &
0.25 &
0.25 \\
\hline
label smoothing & 
0.1 & 
0.1 &
0.1  &
0.1 & 
0.1 & 
0.3 &
0.1  &
0.3 \\
drop path rate & 
0.1 & 
0.2 & 
0.1 & 
0.2 &
0.1 &
0.3 & 
0.2 & 
0.3 \\
\hline
\end{tabular}
}
    \label{tab:supp_cls_22k}
\end{table*}

\noindent\textbf{COCO object detection.} For fair comparisons, we follow common practices~\cite{liu2021swin,liu2022convnet} to initialize the backbone with pretrained weights and train the models using a 3$\times$ (36 epochs) schedule by default. The shorter side is resized to 480$-$800 pixels, while the longer side does not exceed 1,333 pixels. All the models are trained with a batch size of 16 and AdamW~\cite{loshchilov2017decoupled} optimizer with an initial learning rate of $1\times10^{-4}$. We have publicly released the training configuration files used in the MMDetection framework and trained weights.

\noindent\textbf{ADE20K semantic segmentation.} We evaluate UniRepLKNet models on the ADE20K dataset~\cite{zhou2017scene}, and initialize them with the pre-trained classification weights. The learning rate is initialized with $1\times10^{-4}$ and decayed with the polynomial decay schedule with a power of 1.0. Following previous methods~\cite{liu2021swin,liu2022convnet}, the crop size is set to 512 for the ImageNet-1K-pretrained models, and 640 for ImageNet-22K-pretrained models. All segmentation models are trained with a batch size of 16 for 160k iterations. We have publicly released the training configuration files used in the MMSegmentation framework and trained weights.

\section*{Appendix C: Shape Bias}

\begin{figure}
		\includegraphics[width=\linewidth]{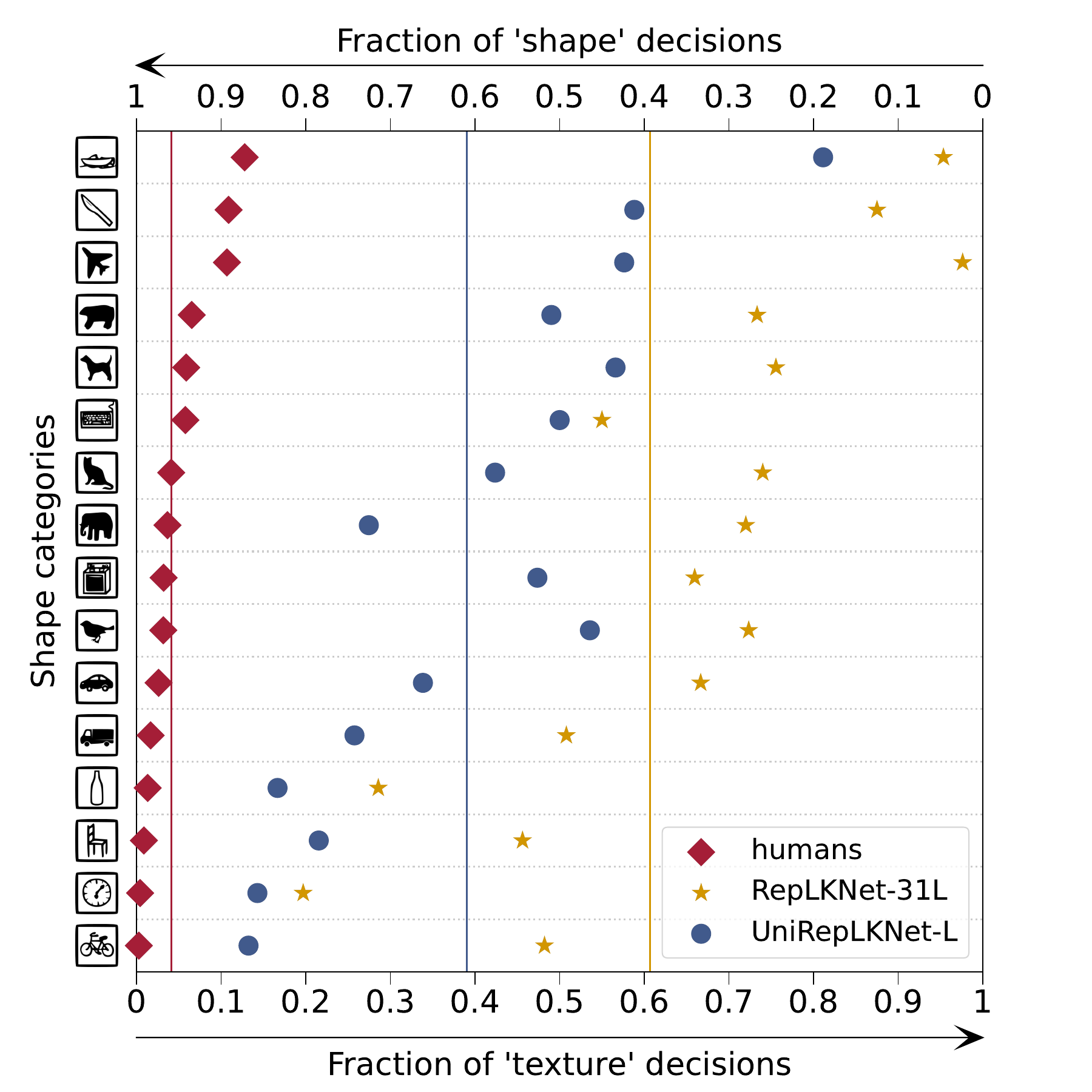}
  \vspace{-0.2in}
		\caption{Shape bias of ImageNet-22K-pretrained UniRepLKNet-L and RepLKNet-31L.}
  \label{fig-shape-bias-unireplknet}
\end{figure}

\begin{figure}
		\includegraphics[width=\linewidth]{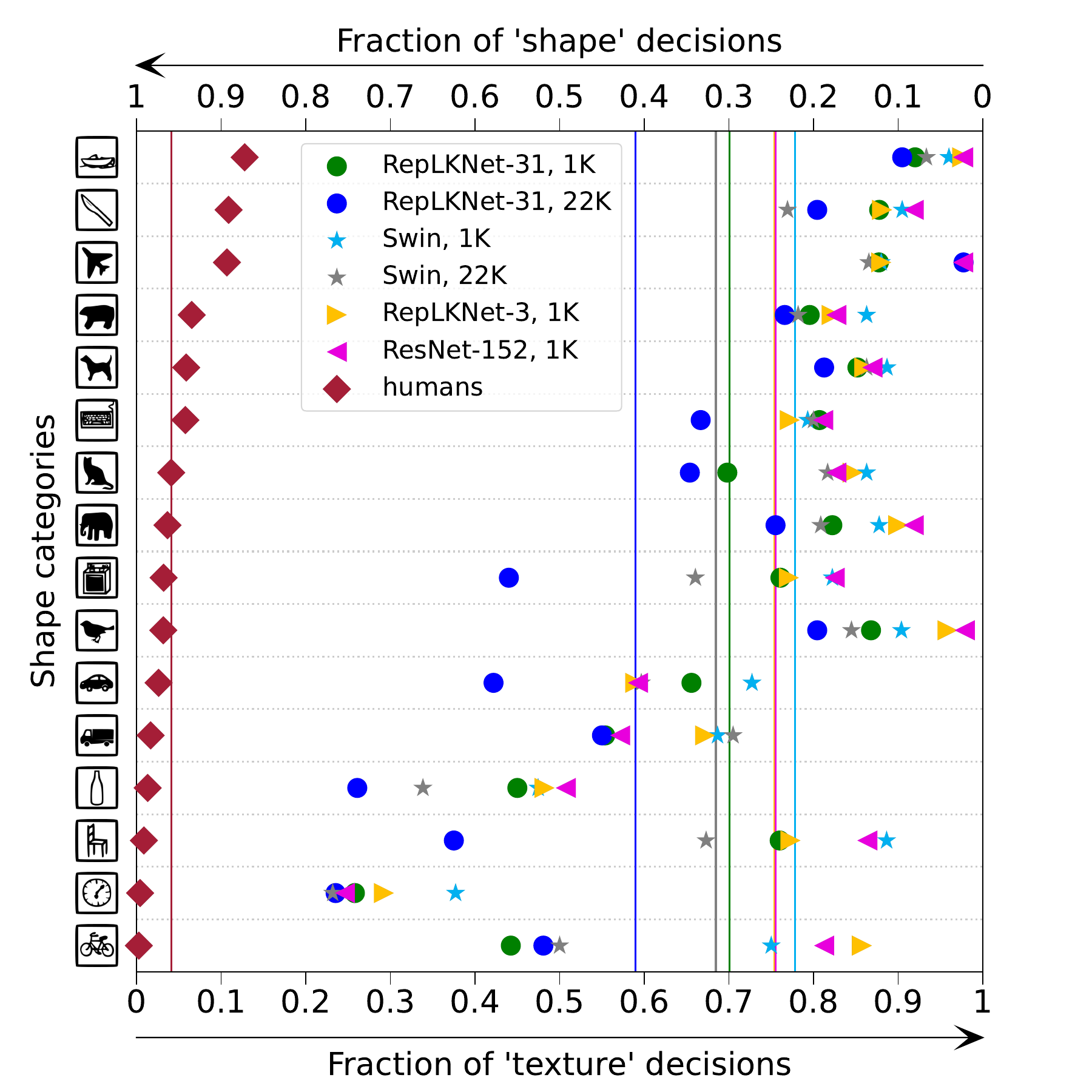}
  \vspace{-0.2in}
		\caption{Shape bias of ImageNet-1K and ImageNet-22K-pretrained RepLKNet-31B and Swin-B. This figure is directly taken from the supplementary material of RepLKNet without any modifications}
  \label{fig-shape-bias-replknet}
\end{figure}

A higher shape bias means the model makes predictions based more on the shape of objects rather than the textures, \ie, the model behaves more similarly to humans. Therefore, a model with a higher shape bias may transfer better to downstream tasks. UniRepLKNet demonstrates significantly higher shape bias than existing ConvNets and ViTs. Concretely, we test the shape bias of ImageNet-22K-pretrained UniRepLKNet-L and RepLKNet-L with the \textit{modelvshuman} toolbox~\footnote{\url{https://github.com/bethgelab/model-vs-human}}. Fig.~\ref{fig-shape-bias-unireplknet} shows a significantly higher shape bias of UniRepLKNet - UniRepLKNet makes 20\% more decisions based on the overall shapes of objects. This improvement is particularly remarkable since RepLKNet is already known to have a high shape bias (Fig.~\ref{fig-shape-bias-replknet} is directly taken from the supplementary material of the RepLKNet paper without any modifications).

\subsection{Appendix D: Training Memory Footprint}

The extra parallel dilated branches in Dilated Reparam Block consume more training resources, which is acceptable considering the performance improvements. We present the peak GPU memory footprint and training speed in Table~\ref{table-costs}. With a bigger model and bigger data, we may trade the performance for higher training speed and lower memory consumption by replacing the Dilated Reparam Block with a single large-kernel conv layer followed by Batch Normalization layer. We test the peak memory footprint and actual training throughput while training UniRepLKNet-S with 224$\times$224 inputs and a batch size of 4096 on a node with eight A100 GPUs. Note that such results are significantly influenced by the hardware environment and specific implementation; thus, they should be considered as references only.

	\begin{table}
		\caption{Training costs.}
		\label{table-costs}
		\vspace{-0.3in}
		\begin{center}
        \resizebox{1.0\linewidth}{!}{
			\tiny
			\begin{tabular}{lcccccccc}
				\hline
			 & Peak memory     & Training throughput \\
				\hline
                Dilated Reparam Block   & 24.6GB   &  6642 images/s   \\
                Single large-kernel conv layer       & 20.8GB    & 9675 images/s \\

				\hline
			\end{tabular}
        }
		\end{center}
		\vspace{-0.3in}
	\end{table}

\end{document}